\documentclass[runningheads,a4paper]{llncs}

\usepackage[flushleft]{threeparttable}
\usepackage{graphicx}
\usepackage{amssymb}
\usepackage{amsmath}
\usepackage{float}
\usepackage{mathtools}
\usepackage{multirow}
\usepackage{array}
\newcolumntype{P}[1]{>{\centering\arraybackslash}p{#1}}
\usepackage{booktabs,caption,fixltx2e}
\usepackage{float}

\begin{document}

\mainmatter  

\title{Rate of Change Analysis for Interestingness Measures}


%
%
\author{Nandan Sudarsanam\inst{1}
\and Nishanth Kumar \inst{2}
\and Abhishek Sharma \inst{3}
\and Balaraman Ravindran \inst{4} }
%

\institute{
	Department of Management Studies, Indian Institute of Technology-Madras\\
	\email{nandan@iitm.ac.in}
	\and IFMR Finance Foundation\\
	\email{Nishanth.K@IFMR.co.in}
	\and Department of Metallurgical and Materials Engineering, Indian Institute of Technology-Madras\\
	\email{shaabhishek@gmail.com}
	\and Department of Computer Science, Indian Institute of Technology-Madras\\
	\email{ravi@cse.iitm.ac.in}
}

%
%

\toctitle{Lecture Notes in Computer Science}
\tocauthor{Authors' Instructions}
\maketitle

\begin{abstract}
The use of Association Rule Mining techniques in diverse contexts and domains has resulted in the creation of numerous interestingness measures. This, in-turn, has motivated researchers to come up with various classification schemes for these measures. One popular approach to classify the objective measures is to assess the set of mathematical properties they satisfy in order to help practitioners select the right measure for a given problem. In this research, we discuss the insufficiency of the existing properties in literature to capture certain behaviors of interestingness measures. This motivates us to present a novel approach to analyze and classify measures. We refer to this as a rate of change analysis (RCA). In this analysis a measure is described by how it varies if there is a unit change in the frequency count $(f_{11},f_{10},f_{01},f_{00})$, for different pre-existing states of the frequency counts. More formally, we look at the first partial derivative of the measure with respect to the various frequency count variables. We then use this analysis to define two new properties, Unit-Null Asymptotic Invariance (UNAI) and Unit-Null Zero Rate (UNZR). UNAI looks at the asymptotic effect of adding frequency patterns, while UNZR looks at the initial effect of adding frequency patterns when they do not pre-exist in the dataset.  We present a comprehensive analysis of 50 interestingness measures and classify them in accordance to the two properties. We also present empirical studies, involving both synthetic and real-world data sets, which are used to cluster various measures according to the rule ranking patterns of the measures. The study concludes with the observation that classification of measures using the empirical clusters share significant similarities to the classification of measures done through the properties presented in this research.

\keywords{ Association rule mining, objective measures, properties of measures, Rate of change analysis}

\end{abstract}

\section{Introduction}

Association rule mining (ARM) has emerged as a powerful and specialized tool to identify patterns in large datasets. It can be used in applications or business operations where instances of some spatio-temporal occurrence is represented in tabular format across a set of common attributes. An ARM study typically results in rules of the form A$\rightarrow$ B, which would mean that, based on evidence from the data, the presence of attribute A is likely to indicate the presence of attribute B. There are two major challenges to an ARM implementation:
 (i) Candidate Generation: This involves the process of filtering all the possible combinations of items that satisfy a given condition for selection. Given the exponentially large possibilities of rules, this condition focuses on the use of frequency based thresholds to remove potentially uninteresting rules \cite{AG93}. 
 The second major challenge is (ii) Candidate Evaluation: This involves the use of an appropriate metric (\textit{interestingness measure}) to evaluate all the different rules that can be defined from the selected item sets \cite{TK}.

This research concerns itself with the latter challenge. Candidate evaluation can be challenging because there are different ways of describing interestingness of rules. A recent study \cite{all} showed that even among \textit{objective measures}, there exist more than 61 that are defined in literature. Also, the information derived from these different interestingness measures (IM) may not always be consistent \cite{TK}. 

The properties are typically defined using a contingency table (see Table \ref{tbl1}), a simplified adaptation from \cite{TK}. Here, two states, present and absent, are defined for two variables, A (rows) and B (columns). The frequency counts $f_{11}$ and $f_{00}$ define the co-presence and co-absence of A and B, respectively. While the term $f_{10}$ would represent the presence of A and absence of B, and $f_{01}$ the opposite.


\begin{table}[h]
	
	\caption{Standard 2x2 contingency table representing the frequency counts of A and B}\label{tbl1}
	\centering
	\begin{tabular}{|c|c|c|}
		\hline \rule[-2ex]{0pt}{5.5ex}  & $B$ & $B^{c}$ \\
		\hline \rule[-2ex]{0pt}{5.5ex} $A$ & $f_{11}$ & $f_{10}$\\  
		\hline \rule[-2ex]{0pt}{5.5ex} $A^{c}$ & $f_{01}$ & $f_{00}$\\   
		\hline 
	\end{tabular} 
\end{table}

In this research, we posit that the popularly used set of 8 properties covered in \cite{TK2} do not fully capture some important aspects of interestingness measures, and this motivates us to define a more relevant and new property based analysis of IMs. Specifically, our motivation is built on the observations of \cite{all}, who state that the 
empirical classification of measures based on how they rank rules has little to do with the property based classification. A deeper study on this mismatch leads us to believe that pre-existing mathematical properties are only useful in specific environmental contexts. These observations lead us to devise simpler, more generic property definitions which can be applied to different environmental contexts and bear a stronger affiliation to rule ranking patterns exhibited by the measures on empirical datasets.

To this end, we create a property definition framework that defines properties based on the change in the IM per unit change in a frequency count $(f_{11},f_{10},f_{01},f_{00})$.
We broadly refer to this as \textit{Rate of Change Analysis} (RCA)\footnote[1]{While this term is used in stock market analysis, our use of this term in the data mining context is novel.}. 
Specifically, we define two properties which look at the partial derivative of the measure at two different pre-existing states of the frequency count. 
The first studies the rate of change behavior of IM when the frequency count is very large (asymptotic effect as the frequency count tends to $+\infty$). We refer to this as \textit{Unit-Null Asymptotic Invariance} (UNAI).
The second property is defined at the point when the frequency count is currently $0$ or is tending to $0$. We refer to this as \textit{Unit-Null Zero Rate} (UNZR). This looks at the effect of increasing the frequency count on the IM when it is currently non-existent in the data set.  
By defining properties based on how measures actually change at different contingency table configurations we explicitly link the rule ranking behavior with the mathematical property.

\subsection{Intuition for the properties UNAI and UNZR}

When UNAI is satisfied, we can say that the measure will not keep increasing or decreasing with the addition of one of the $f_{ij}$s while the others are kept constant, that is, the metric will asymptotically converge to a fixed value. A metric that fails this property will not converge to a constant value with continued addition of $f_{ij}$s. An example is Lift, which keeps increasing with addition of $f_{00}$s and does not converge to a value.

UNZR is satisfied when we can say that the measure will increase when shown evidence of co-presence or co-absence, if such evidence did not previously exist. Also, it should decrease when shown evidence that one item occurs when the other does not (case of counterexamples). Such a relationship could be weak, but at the very least, such metrics will not behave counter to expectation (like decreasing when shown evidence of co-presence or co-absence) and will not stay completely invariant.\\

The major contributions of this research are listed as follows:
\begin{itemize}
	\item Introduction of a novel approach to classify interestingness measures and the development of two specific properties, namely UNAI and UNZR, using this approach
	\item An analysis of the performance of these properties through the classification of various interestingness measures, as well as a comparison with other properties presented in \cite{TK}
	\item Presenting empirical case studies that provide validation for the findings and also demonstrates the usefulness of the properties using real-world and synthetic data sets.
\end{itemize}


\section{Related Work}
\label{sec:prevwork}

A large number of objective IMs have emerged as a result of the application of ARM across different domains. It is also documented that not all measures are capable of capturing the strength of associations and in some cases provide conflicting information of the strength of patterns \cite{TK}. 
Given the abundance of measures and difficulty in choosing the appropriate IM, researchers have suggested various classification schemes (of the IMs) to help identify the appropriate measure for a given application \cite{PS}, \cite{TK}, \cite{TK2}, \cite{GGDN}, \cite{all}, \cite{TGTB}. 
There are two different types of classification that exist in literature: classification based on the properties of IMs (e.g. \cite{PS}, \cite{TK}, \cite{TK2}, \cite{GGDN}) and classification based on empirical results of IMs on different datasets (e.g. \cite{all}). 

Research conducted by \cite{PS} formalized a framework consisting of three properties that an IM should satisfy, namely: the measure should take value 0 if the occurrences of itemsets are independent (P1); the measure should be monotonically increasing with the co-presence of itemsets (P2); and the measure should be monotonically decreasing with the occurrences of either itemsets (P3).

\cite{TK} proposed the following 5 properties in addition to the 3 proposed by \cite{PS}: symmetry under variable permutation (O1), row/column scaling invariance (O2), anti-symmetry under row/column permutation (O3), inversion invariance (O4) and null invariance (O5). They conducted a comparative study, testing 21 different IMs against the resulting 8 properties. The authors further proposed that the optimal way of finding a suitable IM would be to let the user define a property vector indicating the properties that would be ideally required for the given application. This property vector would then be compared to the property vectors of the different objective measure to pick out the ideal interestingness measure for that particular case. For instance, the null-invariance property is considered to be important for interestingness measures used in the context of small probability events in a large dataset \cite{Wu}.
While there has been further work in introducing new properties (e.g., \cite{HH}, \cite{FR}, \cite{GH}, \cite{GGDN}, \cite{Hebert2007}), these have not been as commonly used or cited as the work of \cite{PS} and \cite{TK2}.

There has been limited work on classification of IMs based on empirical results  on different datasets. Research by \cite{Huyn1} proposed the classification of 35 different interestingness measures based on their empirical performance on 2 different datasets by studying the correlation of the interestingness measures. These measures were classified using a graph based clustering approach to create high correlation and low-correlation graphs. 
The work of \cite{all} performed a comprehensive classification of 61 different objective IMs on the based on empirical results on 110 different datasets. It suggested that there exist 21 clusters of measures which are distinct and each of these clusters were studied in detail.

\section{Mathematical definitions for properties UNAI and UNZR}
\label{sec:mathreq}

An interestingness measure (IM) can be represented as a function of the frequency counts (see Equation \ref{eqn:1}).
RCA analysis seeks to assess the relative change in the interestingness measure per unit change of the frequency counts. This is essentially the first partial derivative of the interestingness measure with respect to the variables representing the counts, as shown in Equation \ref{eqn:2}. The set of formulas representing the first partial derivative of the interestingness measure with respect to each of the four state variables  $f_{11}$, $f_{00}$, $f_{10}$ and $f_{01}$ represent the RCA analysis as shown in Equation \ref{eqn:3}. 

\begin{equation}
	\label{eqn:1}
	IM = \phi(f_{11},f_{10},f_{01},f_{00})
\end{equation}
\begin{equation}
	\label{eqn:2}
	\phi^{'}_{f_{ij}} =\frac{\partial(IM)}{\partial f_{ij}}
\end{equation}
\begin{equation}
	\label{eqn:3}
	RCA (IM) = \{\phi^{'}_{f_{11}},\phi^{'}_{f_{10}},\phi^{'}_{f_{01}},\phi^{'}_{f_{00}}\}
\end{equation}
\begin{equation}
\label{eqn:4}
 UNAI_{ij} =  \lim_{f_{ij}\longrightarrow +\infty} (\phi^{'}_{f_{ij}}) 
\end{equation}
\begin{equation}
\label{eqn:5}
UNZR_{ij} =  \lim_{f_{ij}\longrightarrow 0} (\phi^{'}_{f_{ij}}) 
\end{equation}

We use the RCA analysis to define two novel properties.  The \textit{Unit-Null Asymptotic Invariance} (UNAI), and the \textit{Unit-Null Zero Rate} (UNZR).  Mathematically, both these properties are the \textit{derivative at a point} or the \textit{instantaneous rate of change}, at two specific points. We can define the property Unit-Null Asymptotic Invariance (UNAI) as the derivative of the interestingness measure (IM) with respect to $f_{ij}$ as $f_{ij} \to \infty$, and this instantaneous rate of change can be written as shown in Equation \ref{eqn:4}. UNAI can be defined for each of the four frequency count variables by substituting $ij$ with the count of interest. Similar to UNAI, UNZR can be captured by looking at the instantaneous rate of change at $0$. Formally, this would be the derivative of the interestingness measure (IM) with respect to $f_{ij}$ as $f_{ij} \to 0$, and this instantaneous rate of change can be written as shown in Equation \ref{eqn:5}. To compute, UNAIs and UNZRs, in some cases we can simply take the first partial derivative and directly substitute the point of interest, in other scenarios we use the limit notation for derivative at a point (also shown in Equations \ref{eqn:4} and \ref{eqn:5}). Having defined the framework for computing the satisfaction of UNAIs and UNZRs, in the subsequent sections we define the conditions where an interestingness measure can be said to satisfy these properties. These sections presents a classification scheme for the properties UNAI and UNZR which  are presented at the individual $f_{ij}$ level as well as the metric as a whole


\subsection{UNAI property definition}
\label{sub:UNAI}
We create a two-pronged classification scheme for UNAI. We define $UNAI_{f_{ij}}$ which is $UNAI$ defined for each frequency count $(f_{11},f_{10},f_{01},f_{00})$. We do this explicitly for $f_{11}$ which can then be extended to the other frequency counts. We also consolidate the results across all $f_{ij}$s to present the property $UNAI$ for the metric as a whole:
\begin{enumerate} 
	\item $UNAI_{f_{11}}$ is satisfied when: $\lim_{f_{11}\to +\infty} (\phi^{'}_{f_{11}})=0$, for all feasible combination of values of $f_{00},f_{10}, \text{and} f_{01}$. We define a \textit{feasible combination} of values as ones which enable the calculation of the metric in deterministic forms for a database with non-zero rows. \\
 By extension, we can say that the $UNAI_{f_{11}}$ condition is not met when  $\lim_{f_{11}\to +\infty} (\phi^{'}_{f_{11}})\neq0$, for any feasible combination of values of  $f_{00},f_{10}, \text{and} f_{01}$. \\
	 Similarly, we can define $UNAI_{f_{ij}}$ for the other three frequency counts by swapping the variables accordingly.
	\item $UNAI$ is satisfied when $UNAI_{f_{ij}}$ is satisfied $ \forall (ij)$. This is essentially an extension of the classification from $UNAI_{f_{ij}}$ to a general property for the metric as a whole. 
\end{enumerate}

\subsection{UNZR property definition}
\label{sub:UNZR}

The classification scheme we adopt for UNZR is more complex than $UNAI$. Similar to UNAI we adopt a two-pronged approach of defining $UNZR$ at the $f_{ij}$ level as well as a defining it for the metric as a whole. However, we differ from $UNAI$ in that $UNZR$ states are not binary, but have three states that correspond to the property being satisfied, partially satisfied, and not satisfied. Another aspect of the difference is that the definitions at the $f_{ij}$ level are different for \{$f_{11}$, $f_{00}$\} and \{$f_{10}$, $f_{01}$\}. They are identically opposite in terms inequality conditions that need to be met, as shown below. We formally defined the property for $f_{11}$ and $f_{10}$ below and extend it to the other frequency counts $f_{00}$ and $f_{01}$ respectively:
\begin{enumerate}
\item $UNZR_{f_{11}}$ is satisfied when $\lim_{f_{11}\to 0} (\phi^{'}_{f_{11}})>0$ for all feasible combinations of $f_{00},f_{10}, \text{and} f_{01}$. Again, a \textit{feasible combination} is one that enables the computation of the metric in deterministic forms. This formulation can be extended to $UNZR_{f_{00}}$ by swapping the variables accordingly.\\
 $UNZR_{f_{10}}$ is satisfied when $\lim_{f_{10}\to 0} (\phi^{'}_{f_{10}})<0$ for all feasible combinations of $f_{11},f_{00}, \text{and} f_{01}$. This formulation can be extended to $UNZR_{f_{01}}$ by swapping the variables accordingly.

 \item $UNZR_{f_{11}}$ is partially satisfied when two conditions are met. These are: (i) $\lim_{f_{11}\to 0} (\phi^{'}_{f_{11}})\geq 0$ for all feasible combinations of $f_{00},f_{10}, \text{and} f_{01}$, and (ii) $\lim_{f_{11}\to 0} (\phi^{'}_{f_{11}})>0$ for at least one or more feasible combinations of  $f_{00},f_{10},\text{and} f_{01}$. This formulation can be extended to $UNZR_{f_{00}}$ by swapping the variables accordingly.\\
 Similarly, $UNZR_{f_{10}}$ is partially satisfied when two conditions are met. These are: (i) $\lim_{f_{10}\to 0} (\phi^{'}_{f_{10}})\leq 0$ for all feasible combinations of $f_{11},f_{00}, \text{and} f_{01}$, and (ii) $\lim_{f_{10}\to 0} (\phi^{'}_{f_{10}})<0$ for at least one or more feasible combinations of  $f_{11},f_{00},\text{and} f_{01}$. This formulation can be extended to $UNZR_{f_{01}}$ by swapping the variables accordingly.
 
\item Finally, by extension, we can say that $UNZR_{f_{11}}$ is not satisfied when either of these two conditions are met: (i) $\lim_{f_{11}\to 0} (\phi^{'}_{f_{11}})<0$ for any feasible combination of $f_{00},f_{10}, \text{and } f_{01}$ or, (ii) $\lim_{f_{11}\to 0} (\phi^{'}_{f_{11}})=0$ for all feasible combinations of $f_{00},f_{10}, \text{and } f_{01}$. This formulation can be extended to $UNZR_{f_{00}}$ by swapping the variables accordingly.\\
Similarly, we can say that $UNZR_{f_{10}}$ is not satisfied when either of these two conditions are met: (i) $\lim_{f_{10}\to 0} (\phi^{'}_{f_{10}})>0$ for any feasible combination of $f_{11},f_{00}, \text{and } f_{01}$ or, (ii) $\lim_{f_{10}\to 0} (\phi^{'}_{f_{10}})=0$ for all feasible combinations of $f_{11},f_{00}, \text{and } f_{01}$. This formulation can be extended to $UNZR_{f_{01}}$ by swapping the variables accordingly.

\item At the overall metric level we say that $UNZR$ property is satisfied for a metric if the $UNZR_{f_{ij}}$ is satisfied $ \forall (ij)$ . We say that UNZR property is partially satisfied for a metric if $UNZR_{f_{ij}}$ is at least partially satisfied for all  $f_{ij}$s. Finally, a metric fails to satisfy the UNZR property if one or more  $UNZR_{f_{ij}}$s do not satisfy the property. 

\end{enumerate}

\section{Illustrative example of the UNAI and UNZR framework using Lift }
In this sections, we consider the behaviour of the popular interestingness measure, Lift under the UNAI and UNZR properties defined in the previous section. Lift is defined as follows: 
\begin{equation}
	\label{eqn:6}
	Lift(L) = \frac{P(A;B)}{P(A)P(B)} = \frac{f_{11}(f_{11} + f_{01} + f_{10} + f_{00})}{(f_{10} + f_{11})(f_{01}+f_{11})}
\end{equation}

Differentiating w.r.t to $f_{11}$ and simplifying, we get 
\begin{equation}
	\label{eqn:7}
	\frac{\partial(L)}{\partial f_{11}} = \frac{2f_{10}f_{11}f{01} + f_{10}f_{01}(f_{10} + f_{00} +f{01}) - f^2_{11}f_{00}}{(f_{10} +f_{11})^2(f_{01} + f_{11})^2}
\end{equation}
We check the UNAI property for Lift by considering the derivative as $f_{11} \rightarrow \infty$
\begin{equation}
	\label{eqn:8}
	L_{f_{11}}(\infty) = \lim_{f_{11} \longrightarrow \infty} \frac{\partial L}{\partial f_{11}} = 
\lim_{f_{11} \longrightarrow \infty} \frac{2f_{10}f_{11}f{01} + f_{10}f_{01}(f_{10} + f_{00} +f{01}) - f^2_{11}f_{00}}{(f_{10} + f_{11})^2(f_{01} +f_{11})^2}
\end{equation}

After algebraic simplification we can say that the above function is equal to zero for all feasible combinations of $f_{00}$, $f_{10}$ and $f_{01}$. Hence, We can say that Lift satisfies UNAI with respect to $f_{11}$. Similarly, we check for UNAI property with respect to $f_{00}$, $f_{10}$ , $f_{01}$. Hence, We can say that Lift satisfies UNAI with respect to $f_{11}$. Similarly, we check for UNAI property with respect to $f_{00}, f_{10}, f_{01}$.

\begin{equation}
	\label{eqn:9}
	L_{f_{00}}(\infty) = \lim_{f_{00} \longrightarrow \infty} \frac{\partial L}{\partial f_{00}} = \frac{f_{11}}{(f_{01} +f_{11})(f_{10}+f_{11})}
\end{equation}

\begin{equation}
	\label{eqn:10}
	L_{f_{10}}(\infty) = \lim_{f_{10} \longrightarrow \infty} \frac{\partial L}{\partial f_{10}} = 0
\end{equation}

\begin{equation}
	\label{eqn:11}
	L_{f_{01}}(\infty) = \lim_{f_{01} \longrightarrow \infty} \frac{\partial L}{\partial f_{01}} = 0
\end{equation}

Here it is evident that this function is not equal to 0 for all possible values of $ f_{11}, f_{10}, f_{01}$. Hence, we say that $UNAI_{f_{00}}$ is not satisfied but I w.r.t to $UNAI_{f_{11}}, UNAI_{f_{01}}, UNAI_{f_{10}}$ is satisfied.  

We check for the UNZR property for $f_{11}$ by taking the partial derivative at $f_{11}=0$, we get, 
\begin{equation}
	\label{eqn:12}
	L_{f_{11}}(0) = \frac{\partial L}{\partial f_{11}}|_{f_{11}=0} = \frac{f_{10}+f_{00}+f_{01}}{f_{10}f_{01}} 
\end{equation}
Similarly, taking the derivative with respect to $f_{00}, f_{10}, f_{01}$ at 0, we get
\begin{equation}
\label{eqn:13}
	L_{f_{00}}(0) = \frac{\partial L}{\partial f_{00}}|_{f_{00}=0} = \frac{f_{11}}{(f_{11}+f_{10})(f_{11}+f_{01})} 
\end{equation}
\begin{equation}
	\label{eqn:14}
	L_{f_{10}}(0) = \frac{\partial L}{\partial f_{10}}|_{f_{10}=0} = - \frac{(f_{01} + f_{00})}{(f_{11}+f_{01})f_{11}} 
\end{equation}

\begin{equation}
	\label{eqn:15}
	L_{f_{01}}(0) = \frac{\partial L}{\partial f_{01}}|_{f_{01}=0} = - \frac{(f_{10} + f_{00})}{(f_{11}+f_{10})f_{11}} 
\end{equation}

\begin{figure}[t]
\centering
\includegraphics[width=\textwidth]{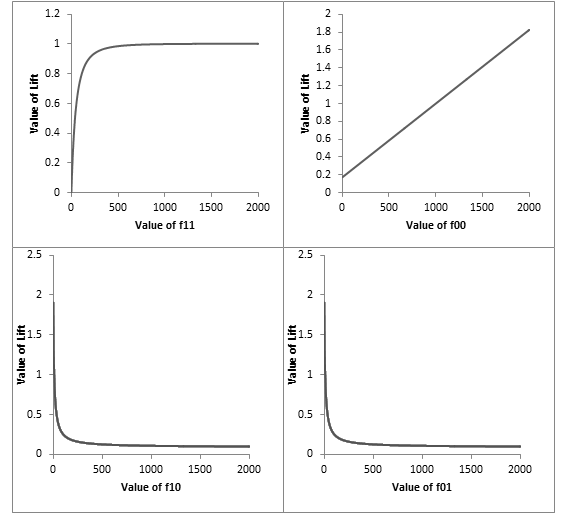}
\caption{Change in value of Lift on varying the frequency counts}
\label{fig:lift}
\end{figure}

We see that for all feasible combinations $UNZR_{f_{11}}$ , $UNZR_{f_{10}}$ and $UNZR_{f_{01}}$ are satisfied. However, $UNZR_{f_{00}}$ is only partially satisfied. From equation \ref{eqn:13} we can see that the following conditions are met: (i) For all feasible combinations of $f_{11}, f_{10}, f_{01}$, $L_{f_{00}}(0) > 0$. This passes the definition of partial satisfaction for UNZR as defined in the paper. At the same time this does not fully satisfy the $ UNZR_{f_{00}}$ property since there are values where it can be 0\footnote{substitute $f_{11}$ = 0, while giving the others positive values}. Figure \ref{fig:lift} 

\section{Mapping UNAI and UNZR to commonly used measures and other properties}
\label{sec:propertyofmetrics}

This section is divided in two parts. The first part performs a detailed analysis that uses the proposed properties to classify commonly used measures. The second part then compares these classifications to the classification done by other popular properties in literature  \cite{TK2}. This two-fold approach is used because it is important to show that a property can actually differentiate between measures (Subsection \ref{subsec4.1}), and that it classifies measures in a way that is different from other properties (Subsection \ref{subsec4.2}).

\subsection{Classification of existing measures using UNAI and UNZR}
\label{subsec4.1}

In this section we classify 50 common measures across the two properties $UNAI$ and $UNZR$, at both the $f_{ij}$ level as well as the metric level. We use all 21 metrics from \cite{TK2} and also borrow popular metrics from \cite{all}. We consciously avoid metrics which are mathematically identical as suggested by \cite{all}, but choose to have metrics which could still be rank-wise indistinguishable. We do this because practitioners might make sense of an absolute score and the rate at which it increases or decreases. We also avoid metrics which need us to make any \textit{a priori} assumptions on probability distributions or cannot be abstracted as a function of $f_{ij}$s. The analysis is carried out in accordance to the definitions in Section \ref{sec:mathreq} and findings are summarized in Table \ref{tbl:maintable}.

\begin{table*}[]
  \centering
	\caption{The $UNAI$ and $UNZR$ properties exhibited by 50 interestingness measures}\label{tbl:maintable}
	\resizebox{\textwidth}{!}
  {\begin{tabular}{l|P{1cm}|P{1cm}|P{1cm}|P{1cm}|P{.7cm}|P{1cm}|P{1cm}|P{1cm}|P{1cm}|P{.7cm}|}
    \textbf{Measure}                                                  & \scalebox{0.75}{$UNAI_{f_{11}}$} & \scalebox{0.75}{$UNAI_{f_{00}}$} &  \scalebox{0.75}{$UNAI_{f_{10}}$}  &  \scalebox{0.75}{$UNAI_{f_{01}}$}  & \scalebox{0.75}{$UNAI$} &  \scalebox{0.75}{$UNZR_{f_{11}}$} & \scalebox{0.75}{$UNZR_{f_{00}}$} & \scalebox{0.75}{$UNZR_{f_{10}}$} & \scalebox{0.75}{$UNZR_{f_{01}}$} & \scalebox{0.75}{$UNZR$}   \\
    \hline

    Lift                                                          & Y       & N       & Y       & Y       & N       & Y       & P       & P       & P       & P  \\
  Jaccard                                                  & Y       & Y       & Y       & Y       & Y       & Y       & N       & P       & P       & N \\
  Confidence                                            & Y       & Y       & Y       & Y       & Y       & Y       & N       & Y       & N       & N \\
  Recall                                                    & Y       & Y       & Y       & Y       & Y       & Y       & N       & N       & Y       & N \\
  Specificity                                             & Y       & Y       & Y       & Y       & Y       & N       & Y       & N       & Y       & N \\
  Precision                                             & Y       & Y       & Y       & Y       & Y       & Y       & N       & Y       & N       & N \\			
  Ganascia                                              & Y       & Y       & Y       & Y       & Y       & Y       & N       & Y       & N       & N  \\
  Kulczynski-1                                         & N       & Y       & Y       & Y       & N       & Y       & N       & P       & P       & N  \\
  F-Measure                                            & Y       & Y       & Y       & Y       & Y       & Y       & N       & P       & P       & N \\
  Causal Confidence                               & Y       & Y       & Y       & Y       & Y       & Y       & Y       & Y       & N       & N \\
  Odd's Ratio                                           & N       & N       & Y       & Y       & N       & P       & P       & P       & P       & P \\
  Negative Reliability                                & Y       & Y       & Y       & Y       & Y       & N       & Y       & N       & Y       & N \\
  Sebag - Schoenauer                             & N       & Y       & Y       & Y       & N       & Y       & N       & P       & N       & N  \\
  Accuracy                                               & Y       & Y       & Y       & Y       & Y       & P       & P       & P       & P       & P  \\
  Support                                                 & Y       & Y       & Y       & Y       & Y       & Y       & N       & P       & P       & N \\
  Coverage                                             & Y       & Y       & Y       & Y       & Y       & P       & N       & N       & P       & N \\
  Prevalence                                          & Y       & Y       & Y       & Y       & Y       & P       & N       & P       & N       & N \\
  Relative Risk                                      & Y       & N       & Y       & Y       & N       & Y       & P       & Y       & P       & P \\
  Novelty                                               & Y       & Y       & Y       & Y       & Y       & Y       & Y       & Y       & Y       & Y \\
  Yule's Q                                             & Y       & Y       & Y       & Y       & Y       & P       & P       & P       & P       & P \\
  Yule's Y                                             & Y       & Y       & Y       & Y       & Y       & P       & P       & P       & P       & P \\
  Cosine                                              & Y       & Y       & Y       & Y       & Y       & Y       & N       & Y       & Y       & N \\
  Least Contradiction                           & Y       & Y       & N       & Y       & N       & Y       & N       & Y       & N       & N \\
  Odd Multiplier                                     & Y       & N       & Y       & Y       & N       & Y       & P       & P       & Y       & P \\
  Descriptive Confirm                            & Y       & Y       & Y       & Y       & Y       & Y       & N       & Y       & N       & N \\
  Causal Confirm                                  & Y       & Y       & Y       & Y       & Y       & Y       & Y       & Y       & N       & N  \\
  Certainty Factor                                 & Y       & Y       & Y       & N       & N       & P       & P       & N       & Y       & N  \\
  Conviction                                         & Y       & Y       & Y       & Y       & Y       & P       & P       & P       & Y       & P \\
  Informational Gain                            & Y       & Y       & Y       & Y       & Y       & Y       & Y       & P       & P       & P \\
  Laplace                                            & Y       & Y       & Y       & Y       & Y       & Y       & N       & Y       & N       & N \\
  Klosgen                                            & Y       & Y       & Y       & Y       & Y       & P       & N       & N       & N       & N \\
  Piatetsky - Shapiro                           & Y       & Y       & Y       & Y       & Y       & Y       & Y       & Y       & Y       & Y  \\
  Zhang                                               & Y       & N       & Y       & N       & N       & Y       & P       & Y       & P       & P \\
  Y and L's 1-way support*                  & Y       & Y       & Y       & Y       & Y       & N       & P       & N       & P       & N \\
  Y and L's 2-way support*                  & Y       & Y       & Y       & Y       & Y       & N       & P       & Y       & Y       & N \\
  Implication Index                              & Y       & Y       & Y       & Y       & Y       & N       & N       & N       & N       & N \\
  Leverage                                         & Y       & Y       & Y       & Y       & Y       & Y       & P       & Y       & N       & N \\
  Kappa                                             & Y       & Y       & Y       & Y       & Y       & P       & P       & Y       & Y       & P \\
  Causal Confirm Confidence            & Y       & Y       & Y       & Y       & Y       & Y       & Y       & Y       & N       & N  \\
  Examples and Counter Examples   & Y       & Y       & N       & Y       & N       & P       & N       & Y       & N       & N \\
  Putative Casual Dependency           & Y       & Y       & Y       & Y       & Y       & P       & P       & Y       & Y       & P \\
  Dependency                                     & Y       & Y       & Y       & Y       & Y       & P       & P       & P       & P       & P \\
  J-measure                                        & Y       & Y       & Y       & Y       & Y       & N       & N       & Y       & N       & N \\
  Collective Strength                           & Y       & Y       & Y       & Y       & Y       & Y       & Y       & Y       & Y       & Y \\
  Gini Index                                            & Y       & Y       & Y       & Y       & Y       & N       & N       & P       & P       & N \\
  Goodman-Kruskal                            & N       & N       & N       & N       & N       & N       & N       & N       & N       & N  \\
  Mutual Information                           & Y       & Y       & Y       & Y       & Y       & N       & N       & Y       & Y       & N \\
  Normalized Mutual Information       & Y       & Y       & Y       & Y       & Y       & N       & N       & N       & N       & N \\
  Loevinger       				         & Y       & Y       & Y       & N       & N       & P       & P       & N       & Y       & N \\
  Added value                                    & N       & Y       & N       & N       & N       & P       & P       & P       & P       & P  
  \end{tabular}
  }
  \par

  \begin{tablenotes}
    \small
    \item 
    \item  Where, Y: Indicates that the Property is Satisfied, P: Indicates that the property is partially satisfied, and N: Indicates that the property is not satisfied\\
    * These metric names are shortened to fit into the table: Y and L's stand for Yao and Liu's for both the shortened names\
  \end{tablenotes}
\end{table*}

The results on the classification of these measures provide two important insights. First, that $UNAI$ property for the metrics as a whole is satisfied by a majority of the measures (37 of the 50). These numbers are even higher for the individual $UNAI_{f_{ij}}$ (ranging from 45 for $f_{11}$, 44 for $f_{00}$, 46 for $f_{10}$ and 45 for $f_{01}$ out of the 50 measures). This suggests that UNAI would be less useful as a tool to eliminate measures that nullify the unstable effect of one frequency count being particularly large. Instead, this property can be useful when due importance needs to be given when a frequency count is expected to be high and continues to grow. A classic scenario would be Lift. In certain contexts, an increase in co-absence in a sparse database should continue to increase the metric value since it makes co-presence even less probabilistic through random chance.\\
The second insight from the case of $UNZR$ is of a different nature. At the overall metric level, there are only 3 measures that fully satisfy the UNZR property, they are \textit{Novelty}, \textit{Piatetsky-Shapiro}  and \textit{Collective Strength}. Of the remaining, 14 measures partially satisfy the property and 33 fail to satisfy the property.  For each $f_{ij}$ the UNZR measures are more discerning. In the case of $f_{11}$, 25 satisfy the property, 9 for $f_{00}$, 22 for$f_{10}$ and 15 for $f_{01}$. These suggest that UNZR at the  $f_{ij}$ level could be more meaningfully used to pick metrics, especially for the case of $f_{00}$, which is satisfied by only nine measures. A particular case could be when the practitioner expects an $f_{ij}$ to be low or close to zero and would like to see the metric impacted when presented with evidence of it. The use of $UNZR$ at the overall metric level could also be useful if the practitioner suspects that any of the frequency values can be close to zero but would like to see its presence or absence to have a meaningful impact on the metric. 

\subsection{Comparing the UNAI and UNZR mapping with other properties}
\label{subsec4.2}
In this section we compare the classification of measures done through $UNZR$ and $UNAI$, with the classification done through other properties in literature \cite{TK2}. This is important because, in addition to fulfilling other criteria, it is necessary that a property classifies measures differently from other pre-existing properties. Otherwise, there is a redundancy and one could question the need for the new property in question. We conduct our comparison on the properties proposed by \cite{TK2}. This includes five new properties proposed in that study, as well as three previous properties from \cite{PS}. In order to perform the analysis, we take all the 50 measures analyzed in Table \ref{tbl:maintable} which include the 21 measures analyzed by \cite{TK2}. We conduct an analysis that compares the classification of these measures across the  two states of $UNAI$ and three states of $UNZR$  and compare it to the two states (satisfied or not satisfied) across the 8 properties presented in \cite{TK2}. This leads us to create the Contingency Table \ref{tbl:table3}.

\begin{table*}[h]
	\centering
	\caption{Contingency Table: Relationship between classification of UNAI and UNZR and the classification of other prominent properties}
	\label{tbl:table3}
	\resizebox{\textwidth}{!}
	{
		\begin{tabular}{@{}llcc|ccc@{}}
			\midrule
			&               &  \multicolumn{2}{c|}{UNAI}           & \multicolumn{3}{c}{UNZR}                                     \\
			 \midrule
			&               & Satisfied & Not Satisfied          & Satisfied              & Partially Satisfied & Not Satisfied \\
			
			\multirow{2}{*}{P1: Statistical independence}  & Satisfied     & 15         & 4                      
			& 2                      & 8                   & 9             \\ 
			
			& Not Satisfied & 22         & 9                     
			 & 1                      & 6                   & 24             \\ 
			\hline
			\multirow{2}{*}{P2:(Refer \cite{PS})}  & Satisfied     & 34         & 13                     
			 & 3                      & 14                   & 30             \\ 
			
			& Not Satisfied & 3         & 0
			& 0                      & 0                   & 3             \\ 
			\hline
			\multirow{2}{*}{P3:(Refer \cite{PS})}  & Satisfied     & 27         & 11
			& 3                      & 14                   & 21             \\ 
			
			& Not Satisfied & 10         & 2                      
			& 0                      & 0                   & 12            \\ 
			\hline
			
			\multirow{2}{*}{O1: Symmetry under variable permutation }  & Satisfied     & 13         & 4
			& 3                     & 7                   & 7             \\ 
			
			& Not Satisfied & 24         & 9
			& 0                      & 7                   & 26             \\ 
			\hline
			\multirow{2}{*}{O2: Row and Column Scaling Invariance}  & Satisfied     & 2         & 1                      
			& 0                      & 3                   & 0             \\
			
			& Not Satisfied & 35        & 12                      
			& 3                      & 11                   & 33            \\
			\hline
			\multirow{2}{*}{O3: Antisymmetry row or column permutation}  & Satisfied     & 4         & 0                      
			& 2                      & 2                   & 0             \\
			& Not Satisfied & 33        & 13                      
			& 1                      & 12                   & 33            \\
			\hline
			\multirow{2}{*}{O3': Inversion Invariance} & Satisfied     & 10         & 1                      
			& 3                      & 5                   & 3             \\
			& Not Satisfied & 27         & 12
			& 0                      & 9                   & 30             \\
			\hline
			\multirow{2}{*}{O4: Null Invariance}  & Satisfied     & 8         & 4
			& 0                      & 0                   & 12             \\
			& Not Satisfied & 29        & 9
			& 3                      & 14                   & 21         \\
			    \midrule 
		\end{tabular}
	}
\end{table*}

The findings from Table \ref{tbl:table3} suggest that the classification of measures through $UNAI$ and $UNZR$ are more or less independent of the classification done through all of the eight pre-existing properties. The few cases where we see low overlaps is also easily explainable by the low membership to a certain class and not a relationship between properties (for instance, observe that only  3 of the 50 measures satisfy the 'Row and Column Scaling Invariance' or fully satisfy UNZR). We do not, however, carry out a Chi-Square test to establish independence because in the case of some properties they are explicitly related. For instance, all Null Invariant properties have to fail UNZR by definition. It is therefore not entirely meaningful to perform such an analysis to look at statistical independence. The overarching conclusion from the Table \ref{tbl:table3} is that while some of these properties could be weakly related to each other, there is sufficient independence with pre-existing properties that can justify UNAI and UNZR as two new properties in-terms of classification of measures.

\section{Empirical Studies }
\label{sec:Casestudy}
The work of \cite{all} has established that empirical clustering of measures bears no meaningful relationship to properties presented in \cite{TK2} (which also cover three properties originally presented in \cite{PS}). While the properties UNAI and UNZR have been constructed to intuitively convey a certain mathematical aspect of the measure, an important motivation and therefore requirement in design was that they have a meaningful map to the actual behavior of measures, empirically. Our studies across a wide range of datasets, both synthetic and real suggest that these two properties bear strong relationships with the empirical clusters. More interestingly, we find that the results are substantially more pronounced in certain environmental conditions. Specifically, we find that $UNZR_{f_{11}}$ and  $UNAI_{f_{00}}$ are valuable in sparse datasets, and correspondingly $UNZR_{f_{00}}$ and $UNAI_{f_{11}}$ are better properties to consider in dense data. In the following sections, we do a detailed and illustrative analysis showing how the $UNZR_{f_{11}}$ classification of measures is useful in sparse datasets and $UNZR_{f_{00}}$ is useful in dense datasets. The motivation to choose the $UNZR$ properties over the $UNAI$ is the fact that the $UNZR$ creates groups of more or less equal sizes. For instance, $UNZR_{f_{11}}$ splits the measures with 25 of them satisfying the property, 15 of them partially satisfying it, and 10 of them failing to satisfy the property. Where as with  $UNAI_{f_{00}}$ we see that 44 of the 50 measures satisfy this property. A similar comparison exists between $UNZR_{f_{00}}$ and  $UNAI_{f_{11}}$.

We conduct our empirical studies by first considering synthetic contingency tables that mimic sparse and dense datasets, and in each case we explore further by choosing a real world dataset that is sparse and dense, respectively. Based on the rule ranking of the measures in the two environmental conditions, we then cluster the measures into sets and see how they correlate with the property of interest.

\subsection{Sparse datasets}
\label{subsec5.1}

Sparse datasets are characterized by having a relatively high $f_{00}$ count with respect to $f_{11}$, primarily, and to a lesser extent $f_{10}$, and $f_{01}$. As discussed in the previous section we choose to analyze the effect of the $UNZR_{f_{11}}$ property in this setting.

We mimic the rules from a synthetic dataset using artificially created sets of rules in form of contingency tables. We do this specifically for the sparse settings. We achieve these environments by assigning low values to ${f_{11}}$, high values for ${f_{00}}$, while ${f_{10}}$, ${f_{01}}$ fall in between the two extremes. The ${f_{11}}$, ${f_{00}}$, ${f_{10}}$ and ${f_{01}}$ cells of the tables took the values \{0, 1, 10, 11\}, \{1000, 5000, 10000, 25000, 50000, 75000, 100000\}, \{10, 100, 250, 500, 600, 800, 1000\} and \{10, 100, 250, 500, 600, 800, 1000\} respectively. This resulted in $1372$ unique contingency tables, each representing a rule in a sparse dataset.

For the real world dataset, we chose the fairly popular 'Adult' data set from the UCI Machine Learning archive \cite{UCI1}. This is essentially an extraction from a census database which has demographic and financial information of individuals. This includes features like age, employment, gender, native country, etc. 

In its native format there are a total of 14 features and more than 48,000 records. A detailed discretization and binarization of variables was carried out in conformance to the best practices suggested in \cite{tankumarbook}. These helps us create the transactional table. This table now has a total of 115 features. We confine the analysis to one-to-one rules. We use a basic support based pruning with a threshold close to 0, in order to get a full enumeration of all one-to-one rules but avoid a variable mapping to itself. This results in a total of $13000$ rules. Similar to the \cite{all} we choose a subset of the rules to compare. However, given the unique nature of our problem, unlike \cite{all} we do not randomly select the rules. Instead we choose a subset of rules that are typically encountered in sparse data sets, by selecting cases where $f_{11}$ is lower than $f_{00}$. This results in $764$ rules.

In the next steps we follow the same procedure as \cite{all}. Each rule is evaluated using each measure, and a rank ordering of rules is done for each measure. Using Spearman's rank correlation, we create a matrix of pairwise distances between measures which acts as the adjacency matrix for a complete graph. We create clusters by using a threshold value of 0.8 on the correlation co-efficient. This process naturally creates groups of measures depending on the threshold used. While there are various other graph clustering algorithms that can be implemented, the simplicity of this approach is appealing.

\begin{table}[t]
	\centering
	\caption{Empirical analysis - Sparse dataset}
	\label{tbl:table4}
	{
		\begin{tabular}{lcc|ccc}
		\midrule


		Dataset	& Cluster & Measures & N & P & Y  \\

		\hline
		 &  & 50 & 10 & 15 & 25\\
		\hline


		\multirow{3}{*}{Synthetic} & A & 21 & 0 & 4 & 17  \\

		 & B & 20 & 4 & 9 & 7 \\

		 & C & 9 & 6 & 2 & 1  \\

		\hline


		\multirow{2}{*}{Adult} & A & 36 & 2 & 12 & 22 \\

		 & B & 14 & 8 & 3 & 3  \\

		\end{tabular}
	}
	
\end{table}

Our study finds that there is a significant match between the three property states and the clusters that are formed for both the synthetic and real data sets. However, this is not a perfect overlap. We split the measures into three clusters in the synthetic setting and into two clusters in the 'Adult' dataset's rules. The cluster memberships are shown below:

\textbf{Synthetic dataset}: \textbf{Cluster A}: \{ Recall, Precision, Confidence, Jaccard, F-Measure, Odd's Ratio, Sebag Schoenauer, Support,  Lift, Ganascia, Kulczynski-1, Relative Risk,  Yule's Q, Yule's Y, Cosine, Odd Multiplier, Information Gain, Laplace, Zhang, Leverage, Examples and Counter Examples \}, \textbf{Cluster B}: \{ Specificity, Negative Reliability, Accuracy, Descriptive Confirm, Causal Confirm, Piatetsky-Shapiro, Novelty, Causal Confidence, Certainty Factor, Loevinger, Conviction, Klosgen, 1-Way Support, 2-Way Support, Kappa, Putative Causal Dependency, Causal Confirm Confidence, Added Value, Collective Strength, Dependency \}, \textbf{Cluster C}: \{ Mutual Information, Coverage, Prevalence, Least Contradiction, Normalized Mutual Information, Implication Index, Gini Index, Goodman Kruskal, J-Measure \}

\textbf{'Adult' dataset}: \textbf{Cluster A}: \{ Recall, Precision, Confidence, Jaccard, F-Measure, Odd's Ratio,
Sebag Schoenauer, Support, Causal Confidence, Lift, Ganascia, Kulczynski-1,
Relative Risk, Piatetsky-Shapiro, Novelty, Yule's Q, Yule's Y, Cosine,
Odd Multiplier, Certainty Factor, Loevinger, Conviction, Information Gain,
Laplace, Klosgen, Zhang, 1-Way Support, 2-Way Support, Leverage, Kappa,
Putative Causal Dependency, Examples and Counter Examples, Causal Confirm Confidence, Added Value, Collective Strength, Dependency \}, \textbf{Cluster B}: \{ Mutual Information, Specificity, Negative Reliability, Accuracy, Coverage, Prevalence, Least Contradiction, Descriptive Confirm, Causal Confirm, Normalized Mutual Information, Implication Index, Gini Index, Goodman Kruskal, J-Measure \}

The relationship between empirical cluster memberships and property affiliations is summarized in Table \ref{tbl:table4}. In the synthetic dataset, all of the 21 measures of cluster A satisfy $UNZR_{f_{11}}$, either completely of partially. The split is rather more even in cluster B, but cluster C is dominated by measures which do not satisfy $UNZR_{f_{11}}$. In the 'Adult' dataset, cluster A again overwhelmingly consists of measures which satisfy $UNZR_{f_{11}}$, either partially or completely (34 out of 36), whereas the properties that do not satisfy $UNZR_{f_{11}}$ tend to exist more in cluster B. 

\subsection{Dense datasets}
\label{subsec5.2}

We characterize dense dataset as one which has relatively higher ${f_{11}}$ count compared to ${f_{00}}$ count, primarily, and to a lesser extent $f_{10}$, and $f_{01}$. As discussed earlier, we choose to study the effect of  $UNZR_{f_{00}}$ property in this environment.

The motivation for using synthetic tables is the same as in the sparse case. The values chosen for ${f_{11}}$, ${f_{00}}$, ${f_{10}}$ and ${f_{01}}$ cells are \{1000, 5000, 10000, 25000, 50000, 75000, 100000\}, \{0, 1, 10, 11\}, \{10, 100, 250, 500, 600, 800, 1000\} and \{10, 100, 250, 500, 600, 800, 1000\} respectively. This resulted in $1372$ unique contingency tables.

For the real world dataset, we chose 'Mushroom' data set from the UCI Machine Learning archive \cite{UCI1}. This data set includes descriptions of hypothetical samples corresponding to 23 species of gilled mushrooms in the Agaricus and Lepiota Family. The methodology of rule generation was identical to that of the 'Adult' dataset, with the focus to create rules from a dense environment (as opposed to the sparse environment in the Adult dataset). This process results in in $739$ rules being used for the purpose of rule ranking.

\begin{table}[t]
	  \centering
	  \caption{Empirical analysis - Dense dataset}
	  \label{tbl:table5}
	  {
		\begin{tabular}{lcc|ccc}
		  \midrule
		  
		  
		  Dataset & Cluster & Measures & N	& P & Y \\
		  \hline
		  
		  & & 50 & 23 & 18 & 9 \\
		  \hline
		  
		  
		  \multirow{3}{*}{Synthetic} & A & 24 & 3 & 15 & 6 \\
		  
		  & B & 19 & 14 & 2 & 3 \\
		  
		  & C & 7 & 6 & 1 & 0 \\
		  
		  \hline
		  
		  
		  \multirow{4}{*}{Mushroom} & A & 23 & 2 & 15 & 6 \\
		  
		  & B & 12 & 7 & 3 & 2 \\
		  
		  & C & 12 & 11 & 0 & 1 \\
		  
		  & D & 3 & 3 & 0 & 0 \\
		  
		\end{tabular}
	  }
	  
	\end{table}

The synthetic dataset was split into 3 clusters while the 'Mushroom' dataset was split into 4 clusters. The cluster memberships are shown below:

\textbf{Synthetic dataset:} \textbf{Cluster A:} \{ Recall, Odd's Ratio, Specificity, Negative Reliability, Lift, Coverage, Piatetsky-Shapiro, Novelty, Yule's Q, Yule's Y, Odd Multiplier, Certainty Factor, Loevinger, Conviction, Information Gain, Klosgen, Zhang, 1-Way Support, 2-Way Support, Kappa, Putative Causal Dependency, Added Value, Collective Strength, Dependency \}\textbf{Cluster B:} \{ Precision, Confidence, Jaccard, F-Measure, Sebag Schoenauer, Support, Accuracy, Causal Confidence, Ganascia, Kulczynski-1, Prevalence, Relative Risk, Cosine, Least Contradiction, Descriptive Confirm, Causal Confirm, Laplace, Examples and Counter Examples, Causal Confirm Confidence \}\textbf{Cluster C:} \{ Mutual Information, Normalized Mutual Information, Implication Index, Gini Index, Goodman Kruskal, Leverage, J-Measure \}

\textbf{'Mushroom' dataset:} \textbf{Cluster A}: \{ Recall, Specificity, Negative Reliability, Lift, Piatetsky-Shapiro, Novelty, Yule's Q, Yule's Y, Odd Multiplier, Certainty Factor, Loevinger, Conviction, Information Gain, Klosgen, Zhang, 1-Way Support, 2-Way Support, Leverage, Kappa, Putative Causal Dependency, Added Value, Collective Strength, Dependency \} \textbf{Cluster B:} \{ Mutual Information, Odd's Ratio, Accuracy, Causal Confidence, Prevalence, Relative Risk, Least Contradiction, Descriptive Confirm, Causal Confirm, Normalized Mutual Information, Gini Index, J-Measure \} \textbf{Cluster C:} \{ Precision, Confidence, Jaccard, F-Measure, Sebag Schoenauer, Support, Ganascia, Kulczynski-1, Cosine, Laplace, Examples and Counter Examples, Causal Confirm Confidence \} \textbf{Cluster D:} \{ Coverage, Implication Index, Goodman Kruskal \}

The results from this analysis is summarized in Table \ref{tbl:table5}. In the synthetic dataset, cluster A is populated by measures which satisfy the $UNZR_{f_{00}}$ (21 out of 24), either partially or completely. Clusters B (14 out of 19) and C (6 out of 7) are dominated by measures that do not satisfy $UNZR_{f_{00}}$. In the 'Mushroom' dataset, cluster A is again consisted of measures which satisfy $UNZR_{f_{00}}$, either partially or completely (21 out of 23). Cluster B is split between the measures that satisfy $UNZR_{f_{00}}$ and measure that don't (7 N's vs 3 P's and 2 Y's). Clusters C and D are overwhelmingly consisted of measures which don't satisfy $UNZR_{f_{00}}$, with only 1 measure satisfying the property among the 15 in both clusters combined. In general, it is evident that the clustering holds a clear mapping to the $UNZR_{f_{00}}$ property for the selected rules in a dense setting.

\section{Conclusions and Future work}
\label{sec:Conclusion}
This study presents a new property-based framework (RCA) for analyzing interestingness measures. This framework uses the partial derivative of an IM with respect to a frequency count. This provides us with the insight of how the IM will change when the frequency count is increased or decrease. This approach is then used to create two specific properties, $UNAI$ and $UNZR$, which correspond to taking the partial derivative at two points, infinity and zero. The study then showcases the classification of a broad set of measures in accordance to these properties and also compares them to the classification done by other properties in literature. The properties proposed in this study classify the measures assigning memberships to all property states, suggesting that they might be discerning some meaningful differences in the measures. The classifications through these properties are also fairly independent of those done by other pre-existing properties, suggesting, that something new is being captured. Finally, the study showcases the utility of classification through the new properties by conducting empirical analyses on both synthetic and real-world data sets, which relate the rule ranking behavior of the measures with two of the properties proposed. The findings suggest that the rule ranking behavior holds a clear relationship to the classification done by the property.

One of the major contributions of this research is the new framework (RCA) for analyzing measures using the rate of change idea through partial differentiation. This is markedly different from the property-based classification schemes that currently exist in literature. Given this, we feel that there could be a more extensions in the development of properties that build on this idea, which go beyond the two that are proposed in this study. Also, the idea of using differentiation as tool to defining properties opens up a plethora of characteristics that can be analyzed. One possible extension is to study the shape of the partial derivative curve (linear, polynomial, etc).

Finally, the authors in this study agree with the view put forth in \cite{all} that meaningful classification of measures needs to, also, be driven by similarity (or dissimilarity) in rule ranking that can be seen on empirical data sets.  We would like to extend this argument by stating that the value of mathematical properties, derived from principled arguments, can be benchmarked across-the-board in this fashion (this study performs such an analysis exclusively for the two properties proposed in this study). This can also be extended beyond Interestingness measures in ARM. We can see that classification metrics (some of which are included in this analysis like accuracy, recall, specificity, etc.) can also be defined by the same contingency table (for two class classification problems) and could therefore lend themselves to a representation and segmentation using a rate of change analysis. 

\section*{Acknowledgments}\label{sec:Acknowledgments}
This work was supported by a funding from IIT Madras (CSE/14-15/831/RFTP/BRAV)

\bibliography{Citations}{}
\bibliographystyle{acm}

\end{document}